\documentclass[twocolumn, secnumarabic, amssymb, nobibnotes, superscriptaddress, aps, prl]{revtex4-1}
\usepackage{amsmath,graphicx,float,color}

\begin{document}

\title{An $O(N)$ Sorting Algorithm: Machine Learning Sort}

\author{Hanqing Zhao}
\affiliation{Department of Modern Physics, University of Science and Technology of China, Hefei 230026, China }
\affiliation{School of Physical Science and Technology, and Key Laboratory for Magnetism and Magnetic Materials of MOE, Lanzhou University, Lanzhou, Gansu 730000, China}
\email{phzhq@mail.ustc.edu.cn}

\author{Yuehan Luo}
\affiliation{Department of Applied Mathematics and Statistics, State University of New York at Stony Brook, Stony Brook, NY, 11794, USA}

\begin{abstract}
We propose an $O(N\cdot M)$ sorting algorithm by Machine Learning method, which shows a huge potential sorting big data. 	This sorting algorithm can be applied to parallel sorting and is suitable for GPU or TPU acceleration.  Furthermore, we discuss the application of this algorithm to sparse hash table.
\end{abstract}
\maketitle

\section{Introduction}
Sorting, as a fundamental operation on data, has attracted intensive interests from the beginning of computing~\cite{cormen2009introduction}. Lots of classic algorithms have been designed and applied, such as Bubble Sort, Selection Sort, Insertion Sort, etc. However, it's been proven that sorting algorithms based on comparison have a fundamental requirement of $\Omega(N\log N)$ comparisons\cite{Ford1959A,art}, which implies the time complexity is at least $O(N\log N)$\cite{Hoare1962Quicksort,Floyd1964Algorithm,232}. The non-comparison sorting algorithms, such as Bucket Bort, Counting Sort and Radix Sort, are not restricted by the $\Omega(N\log N)$ boundary, and can reach $O(N)$ complexity \cite{Isaac1956Sorting}, but these algorithms have very limited applications.
Most of the state-of-art sorting algorithms employ parallel computing to handle big datasets and have accomplished outstanding achievements~\cite{helman1998randomized,shi1992parallel,mihhailov2010parallel,amato1996comparison,khan2011analysis}. For example~\cite{benchmark}, in 2015, FuxiSort~\cite{fuxisort}, developed by Alibaba Group, is a distributed sort implementation on top of Apsara. FuxiSort is able to complete the 100TB Daytona GraySort benchmark in 377 seconds on random non-skewed dataset and 510 seconds on skewed dataset, and Indy GraySort benchmark in 329 seconds. Then, in 2016, Tencent Sort~\cite{tencent} has achieved a speed of 60.7 TB/min in sorting 100 TB data for the Indy GraySort, using a cluster of 512 OpenPOWER servers optimized for hyperscale data centers. However, these algorithms are still limited by the lower boundary complexity of comparison sorting algorithm and time-consuming networking~\cite{5009385}.

On the other hand, machine learning is a field that has been developing rapidly these years, and has been widely applied across different areas~\cite{lecun2015deep,zhao2017copy,esteva2017dermatologist}. In 2012, the emergence of ImageNet classification~\cite{NIPS2012_4824} with deep convolutional neural networks was a great breakthrough that almost halve the error rate for object recognition, and precipitated the rapid adoption of deep learning by the computer vision community. In March 2016, AlphaGo~\cite{silver2017mastering} utilized neural network to beat the human world champion Lee Sedol in the game of Go, which was a grand challenge of artificial intelligence (AI). The huge success of machine learning shows that computer AI could go beyond human knowledge in complicated tasks, even starting from scratch. After that, machine learning algorithms had been widely applied to various areas such as human vision, natural language understanding, medical image processing, etc., and had achieved great accomplishments. With the breakthrough of these algorithms, the improvements in hardware support the AI algorithms works more efficient, such as GPU/TPU acceleration.

Neural network models~\cite{gevrey2003review,schmidhuber2015deep}, as an important group of algorithms used for machine learning, are inspired by the biology of human brains. Classic neural network models have input layer, output layer and hidden layers. Hidden layers consist of lots of connecting artificial neurons. These neurons are tuned according to the input and output data, to precisely reflect the relationship. The nature of the neural network is a mapping from the input data to output data. Once the training phase is done, we can apply this neuron network to make a prediction of unknown data. This is the so-called inference phase. The precision and efficiency of inference phase inspires us to apply machine learning skills to sorting, because in some way, we can treat sorting as a mapping from the data to its ranking in the data set.

In this paper, we propose a sorting algorithm with a complexity of $O(N\cdot M)$ using machine learning, which works especially well on big data. Here $M$ is a small constant, indicating the number of neurons in the hidden-layers of neural network. We first use a small training data set to approximate the distribution of the whole data set through a 3-layer neural network~\cite{zhao2017general}, then apply this neural network to predict the rankings of data in the future fully sorted sequence of size $N$. Note that in the inference phase, comparison operations are not needed. After the inference phase is completed, each number is assigned to its estimate ranking position in an almost sorted sequence. According to the monotonicity of our neural network, with an additional $O(N)$ bucket sort operation, the data set can be fully sorted. Furthermore, the realization of this algorithm in sparse hash table is discussed.

\section{Algorithm}
Suppose we have a data sequence $S$ of real numbers, with a size of $N$, upper bound $x_{max}$ and lower bound $x_{min}$. An effective sorting algorithm must make sure the output sequence $S'$ is fully sorted by exchanging the positions in the data sequence $\{x_i\}$. Suppose a real number $x_i$ ranks $r_i$ in $S'$, the sorting problem can be treated as a bijective function $G(\cdot)$, and $G(x_i)=r_i$. If the function $G(\cdot)$ is known in advance, then the complexity of sorting problem is $O(N)$ \cite{art}. In fact, if all the numbers in $S$ come from a probability distribution $f(x)$, when $N$ gets larger and larger, the ranking $r_i$ of $x_i$ in $S'$ approximately equals to
\begin{equation}
G(x_i)=r_i \approx N F(x_i) =N\int_{x_{min}}^{x_i}f(x)dx,
\label{eq1}
\end{equation}
$F(\cdot)$ is the cumulative probability distribution function of the data set $S$. When $N$ tends to infinity, the equality holds.

However, the biggest problem is that the function $G(\cdot)$ is usually hard to obtain, so is the distribution function $f(\cdot)$. When we're dealing with big data set, $N$ is large enough for the sequence to be processed in a statistical way. Hence, if the distribution function $f(\cdot)$ can be obtained by an efficient approach, then we can decrease the complexity of sorting algorithm to $O(N)$ by applying Eq. (\ref{eq1}).

In order to do that, we choose $N_0$ numbers out of sequence $S$ randomly, which is called the training sequence $A$, $A=\{a_1, a_2,\cdots, a_{N_0}\}$. This training sequence $A$ should share a similar probability distribution to the distribution of the whole sequence $S$. $A$ is sorted using the conventional comparison sorting algorithm to get a fully sorted sequence $A^\prime$, consequently $a^\prime_i\ge a^\prime_{i+1}$ for any $a^\prime_i,~ a^\prime_{i+1} \in A^\prime$. Take $a^\prime_i$ as the input that ranks $i$ in $A^\prime$, $i/N_0$ as the output, the correspondence between $i/N_0$ and $a^\prime_i$ can then be considered as a cumulative distribution function $F(x;N_0)$ of sequence $A$. In fact, $F(x;N_0)$ is an approximate function of the distribution function $F(x;N)$ of the whole sequence $S$. Now, the sorting problem has been transformed into a fitting problem of distribution function.

As an example, suppose the fitting function $F(x;N_0)$ is piecewise linear, which means the data are assumed to follow a uniform distribution on the interval $(a^\prime_i, ~ a^\prime_{i+1})$, for any $x_j$ in $S$ and $x_j\in (a^\prime_i, ~a^\prime_{i+1})$, we have
\begin{equation}
r_j=i\cdot \frac{N}{N_0}+\frac{a^\prime_i-x_j}{a^\prime_i-a^\prime_{i+1}}\cdot \frac{N}{N_0},
\end{equation}
where $r_j$ is the estimate ranking of data $x_j$.

However, in experiments we find that, this assumption of piecewise linear fitting is too rough and will cause too much error in the estimation of ranking. For example, when we set $N=10^4,~ N_0=10^3$, where all the data $x$ apply to a normal distribution $ \mathcal{N}(0,1)$, it turned out that the estimate ranking might have a deviation larger than $100$ from its real ranking. Under some other predetermined assumptions of fitting function, the deviation might be smaller, but it's obvious that none of them will be able to fit in all kinds of distributions perfectly. We realize that the key to this problem is to find a good fitting approach.

    \begin{figure}
    \centering
    \includegraphics[width=8.9cm]{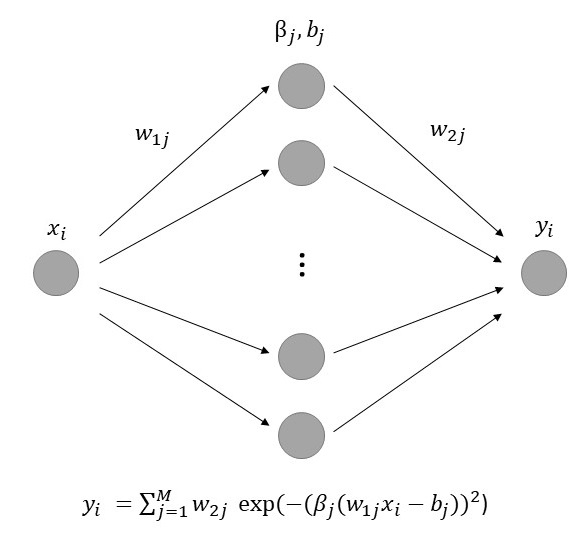}
    \caption{Schematic representation of the GVM. The number of neurons in hidden-layer $M$ is set to 50 in general experiments, while for some distributions that are closer to uniform distribution, $M=10$. }
    \end{figure}

 With artificial neural networks, we can make the fitting function much more precise. In this paper, we apply General Vector Machine (GVM), which is a 3-layer neural network and has only 1 hidden layer. The structure of GVM is schematically shown in the Fig. 1. The learning process of GVM is based on Monte-Carlo algorithm instead of back propagation. We find that GVM is particularly suitable for function fitting \cite{zhao2017general}. This is because when machine learning models are used to predict the ranking of $x_i$, the complexity of machine learning model would strongly affect the complexity of the entire algorithm. The less the number of neurons a machine learning model has and the simpler the structure of the machine learning model is, the more efficient the sorting algorithm is. It still remains to be exploited whether there are other machine learning models that are suitable for sorting problem.

In GVM, the input layer has only one neuron, which is $x_i$ in $S$, and the output layer also has one neuron, which is a real number $y_i$. We fix the number of neurons for the hidden layer to be $M=50$. Specifically, for distributions that are relatively smooth, $M$ could be as small as $10$. $M$ is independent of $N$, and there is no lower bound for $M$ like $\log(N)$, analytically. For example, if the distribution of data sequence is Gaussian, the neural network with only $1$ neuron can accomplish the  work theoretically. In practice, we may have to set more than $1$ neuron to fit in the Gaussian distribution, and the number of neurons depends on the machine learning model. In fact, if the training phase is successful, the more neurons we have in the hidden layer, the more precise our fitting is, but it comes along with the problem of overfitting, as well as decreasing of the computational efficiency. After being trained with the data $(a^\prime_i,~ i)$ in $A^\prime$, GVM can be used as a fitting function to do prediction. Then input the data $x_j$ in $S$, and the neural network will output a real number $y_j$, and $r_j=round(y_j\cdot \frac{N}{N_0})$ is the estimate ranking of $x_j$ in sequence $S$ given by the machine learning model. We round $y_j\cdot \frac{N}{N_0}$ so that $r_j$ is an integer in $[0, N]$. Actually, sometimes $r_j$ would fall out of the range $[0, N]$ because of some small deviation and insufficient sampling, but that doesn't result in a big problem if the size of such kind of data is small enough.

Now we have an estimate ranking for each number $x_j$ in $S$. Though the estimate ranking may have a little deviation from its real ranking, as long as it's in a tolerable range, it's already valuable for some specific applications. Especially when we simply want to know the approximate rankings of  one or several numbers, the complexity of the prediction for each number is only $O(M)$.
After all, the process of estimating the ranking of $N$ numbers takes $O(N\cdot M)$ time.

After the estimate rankings of data are given, to get a fully sorted sequence, the operation of merging these $N$ numbers into a the sequence takes $O(N)$ time. A simple way to do this is to put these $N$ numbers into a linked list of size $N$, and then traverse the list with a comb of size $L$, which means to sort every successive $L$ numbers in the list. Here $L$ depends on how successful the machine learning is. If machine learning process is successful, usually $L$ is a very small integer.

However, based on the property of monotonicity, we propose an approach without combing here. Note that the mapping from a number $x_i$ to the its real ranking $r_i'$ is monotone. Therefore, if the machine learning model is designed to be monotone, we can make sure that, if $x_a\ge x_b$, then $r_a'\le r_b'$. For example, for a three-layer neural network, the function of input $x_i$ and output $y_i$ is
\begin{equation}
y_i=\sum_{j=1}^M w_{2j} \cdot f_0(\beta_j(w_{1j}x_i-b_j)).
\end{equation}
Here $w_{1j}$ and $w_{2j}$ is the weight of each neuron, $b_j$ is bias, $f_0$ is a nonlinear activation function, $\beta_j$ is a multiple factor of GVM for adjusting the sensitivity of nonlinear activation function. In order to guarantee that the machine learning model monotonically decreases, the following condition has to be satisfied:

\begin{align}
d y_i=&\sum_{j=1}^m w_{1j}w_{2j}\beta_j\cdot df_0(\beta_j(w_{1j} x_i-b_j))\le 0,
\label{mono}
\\\nonumber
& \forall x_i \in [-\infty,+\infty].
\end{align}
Normally, we only have to guarantee that Eq.(\ref{mono}) holds for $ \forall x_i \in [x_{min},x_{max}].$

If the machine learning model is designed to be monotone, $N$ numbers are put into their corresponding estimate ranking buckets labeled from $1$ to $N$. The numbers across different ranking buckets must be ordered, but the numbers inside each ranking bucket may not be ordered. Therefore only the numbers inside each bucket need to be sorted. The expected number of numbers inside each bucket is $1$, but because of randomness and unavoidable errors, some buckets may have more than one number, while some other buckets are empty. Suppose the data set applies to a certain distribution, and $N$ numbers are put into $N$ buckets according to their estimate ranking, this operation is actually equivalent to transfering a certain distribution into an uniform distribution. The probability for a bucket to have $q$ numbers is
\begin{equation}
p(x=q)=\binom{N}{q}(\frac{1}{N})^q(1-\frac{1}{N})^{N-q}.
\label{bin}
\end{equation}

We will introduce in the following section that, for a distribution which varies a lot from uniform distribution, the theoretical value of Eq.(\ref{bin}) matches well with the experiment results.

Next we'd like to discuss about the potential combination of parallel computing and our algorithm. Since in the prediction process, it needs no comparison and exchange operations, and the estimation of ranking of each number is independent of each other, it would be efficient to combine with parallel computing, and it requires very little networking workload. In addition to efficient parallel computing, since the machine learning requires matrix manipulation, it's also suitable for executing on GPUs or TPUs for acceleration~\cite{jouppi2017datacenter}.

    \begin{figure}
    \centering
    \includegraphics[width=8.9cm]{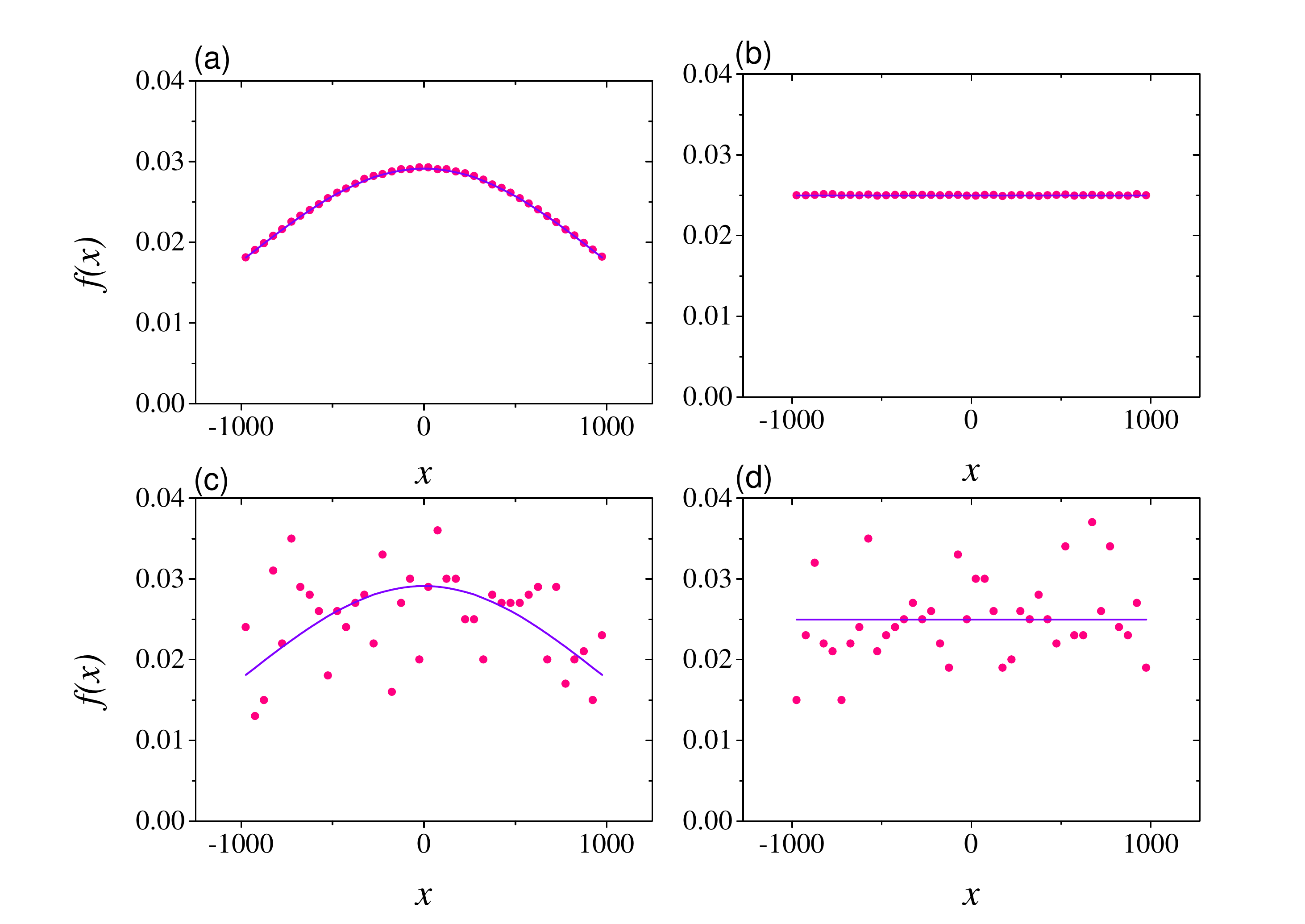}
    \caption{Distribution of the data. The blue lines represent the analytical distribution and pink dots represent the experiment data.  The truncated normal distribution (a) and the uniform distribution (b) for a data set with a size of $10^7$. The truncated normal distribution (c) and the uniform distribution (c) for a data set with a size of $10^3$.  }
    \end{figure}

\section{Experiments}
As shown in Fig. 2, we first run experiments on two kinds of distributions: uniform distribution and truncated normal distribution. The size of data ranges from $N=10^3$ to $N=10^7$. The type of all the data are double, and they are on the interval $[-1000, 1000]$. These two kinds of distributions(Fig. 2) are relatively smooth, their cumulative functions are linear or almost linear. In this experiment, a machine learning model with $M=10$ is used, and the time cost is shown in Fig. 3. The time cost in training machine learning models is excluded here. The time grows with $N$ linearly, and the time cost in each experiment is almost the same. As a baseline, the sorted function in \textit{Python3.6}, takes 10 seconds to sort $10^7$ numbers. The Quick Sort function from \textit{NumPy} package, which is highly optimized, takes 0.9 second to sort $10^7$ numbers, and the Merge Sort from \textit{NumPy} takes 2 seconds to sort $10^7$ numbers. All the experiments above are run using a single CPU, and no further optimization is used in Machine Learning Sort.

    \begin{figure}
    \centering
    \includegraphics[width=8.9cm]{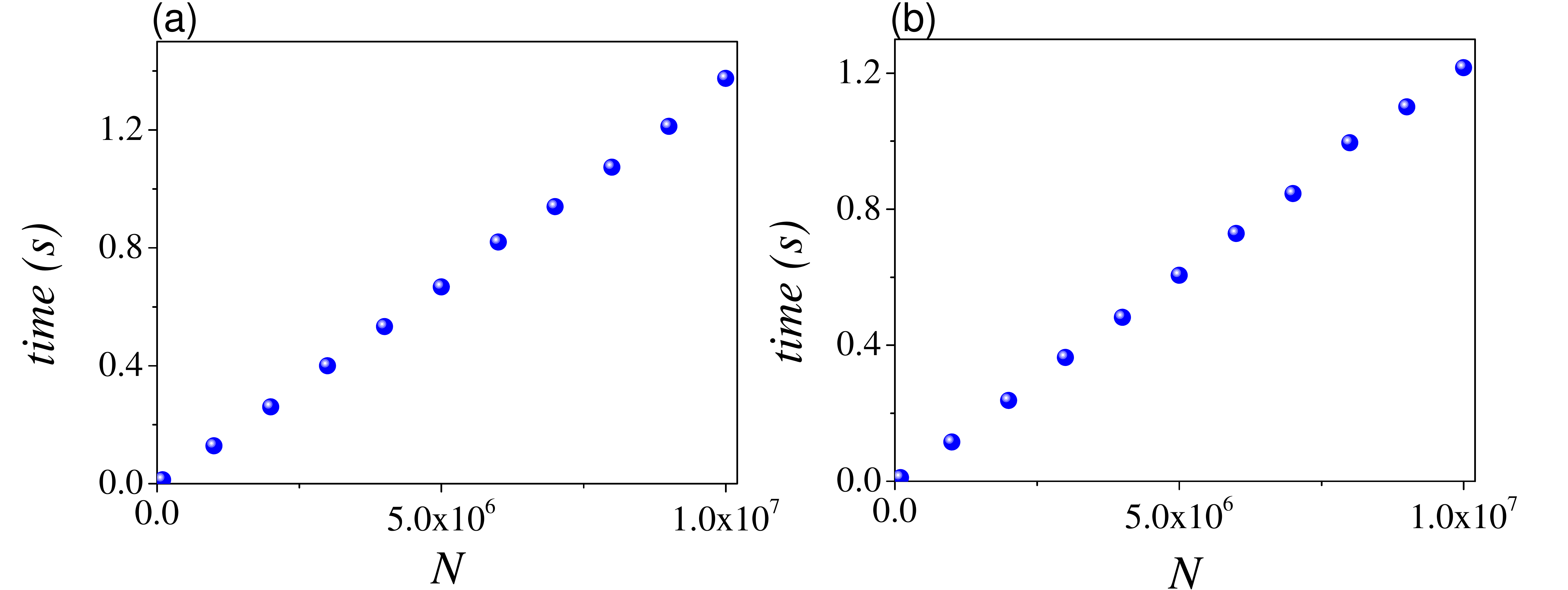}
    \caption{The relation between time complexity and size of data sets for truncated normal distribution (a) and  uniform distribution (b). The blue dots represent the time for the Machine Learning Sort. $10^2$ times of experiments are done for each $N$, and we take the average.}
    \end{figure}

We demonstrate some more examples of more complicated probability distribution functions as shown in Fig. 4(a)-(c), the sorting results are shown in Fig. 4(d)-(f). In each of these experiments, we extract a training data set with a size of $10^4$, and fix the number of neurons to be 50. It turns out that Machine Learning Sort could still fulfill the sorting task perfectly for a data set with a size of $10^7$. Even in the case that the distribution is nothing similar to uniform distribution, the expected size of data in each position is still $1$, and Eq. (\ref{bin}) also holds as Fig. 5 shows.

    \begin{figure}
    \centering
    \includegraphics[width=8.9cm]{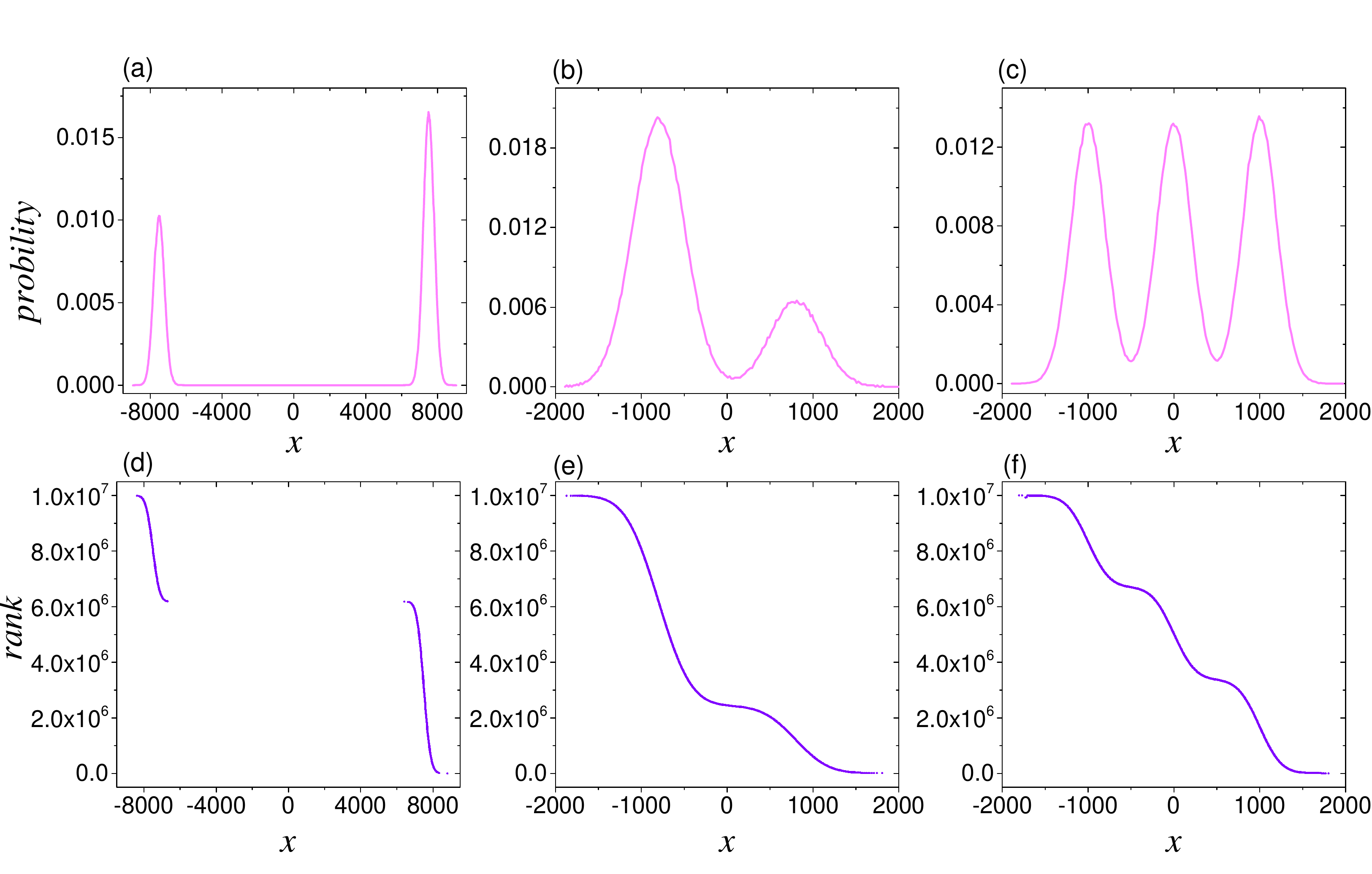}
    \caption{Three different distributions that are not trivial (a)-(c). (d)-(f) shows the sorting results corresponding to (a)-(c), respectively. $N=10^7$.}
    \end{figure}

We should point out that, since the training sequence is only a subset of the whole sequence, the distribution of training sequence can only approximate to the distribution of the whole sequence, for the tail of a distribution (like in Gaussian distribution), the machine learning fitting is not precise enough. However, the tail of a distribution means that only a little fraction of data would fall into that interval. Hence we can gather the data that fall into the tail intervals, and use conventional sorting algorithms (like Quick Sort) to sort them. The size of these data is small enough to have no impact on the overall $O(N)$ complexity. We have to admit that, if there are too many numbers that fall into the tail intervals, the Machine Learning Sort might fail.

    \begin{figure}
    \centering
    \includegraphics[width=8.9cm]{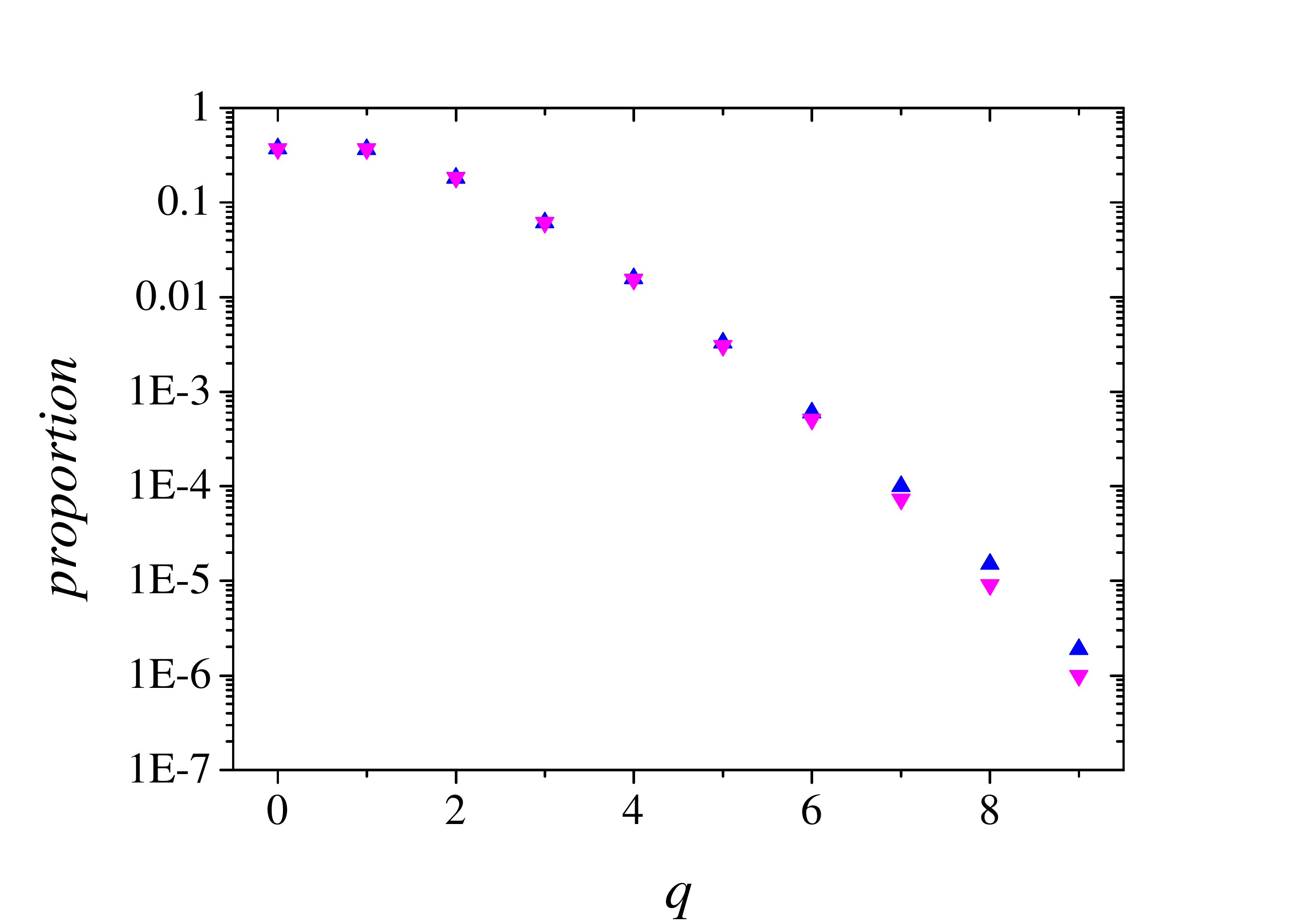}
    \caption{The proportion of the buckets which contains q numbers. The blue triangle represent the experiments' results, and the red triangles represent theoretical values. The origin data follow the distribution in Fig. 4(f), and size of data is $N=10^7$.}
    \end{figure}

\section{Conclusion}
In this paper, we propose a sorting algorithm with complexity of $O(N\cdot M)$. This algorithm is built upon valid machine learning, and it works well for big data sets. That is because the big data sets are usually big enough to apply to certain statistical properties, and their probability distributions are relatively smooth. In the case that the distribution is not easy to learn, to improve the quality of learning machine, we'd like to offer some solutions. First, tune the initial parameters of machine learning model, and pick the best one to proceed to the sorting phase. Second, which is more doable, divide the unsmooth region into relatively smooth region pieces and then apply our algorithm to each piece.
\par
Of course the learning quality also depends on the machine learning models. Here we apply the GVM algorithm. Other machine learning models may also be able to fit in a fair distribution, however, the advantage of GVM is that it has only 3 neuron layers, and is more efficient when fitting functions with few hidden-layer neurons. GVM also has some theories of parameters' tuning. What's more important, for a machine learning model, is that whether it has a structure of monotonicity, which implies that if the sorting is in descendent order and $x_a\ge x_b$, then $r(a)\le r(b)$. This kind of learning model could be applied to sorting algorithm naturally.
\par
Compared to the conventional sorting algorithms, Machine Learning Sort has a better potential in parallel computing. Though the complexity is $O(N\cdot M)$, both $N$ and $M$ can be separated for parallel computing. Unlike the conventional sorting algorithms which mainly utilize a large number of comparison operations, a large portion of operations in Machine Learning Sort is multiplication of real numbers, which can be done efficiently by matrix operation, and is perfect for GPU and TPU acceleration.
\par
This algorithm can also be applied to sparse hash table to reduce space complexity. Use Machine Learning Sort to completely sort the data first, and then set the data and its ranking as the training data set, so that the key code is its ranking. When searching for new data, we put the data into the GVM and get its estimate ranking. The real ranking of the new data is around the estimate ranking with a small deviation.
\par
For a lot of online web companies, they have a need for highly efficient management of big data sets. The statistical properties of these data sets would not vary too much in a short period of time, so we don't have to tune the parameters each time when applying machine learning sort algorithm. For some long-term running models, if the distribution changes slowly over time, we could also consider a machine learning model that changes over time to fit in the changing data. Also, for some applications which only need an approximate ranking instead of fully sorted data sequence, the learning machine would be very suitable.

\begin{acknowledgments}
We are grateful to Yongqing Liang for useful discussions.
\end{acknowledgments}

\bibliographystyle{apsrev4-1}
\bibliography{NEW}

\end{document}